\documentclass{article}
\usepackage{arxiv}

\usepackage[backend=biber, style=authoryear, url=true, doi=false, eprint=false]{biblatex}
\let\cite\parencite  

\usepackage[utf8]{inputenc} 
\usepackage[T1]{fontenc}    
\usepackage{hyperref}       
\usepackage{url}            
\usepackage{booktabs}       
\usepackage{amsfonts}       
\usepackage{nicefrac}       
\usepackage{microtype}      
\usepackage{lipsum}
\usepackage{graphicx}
\graphicspath{ {./images/} }

\title{Improved Emotional Alignment of AI and Humans: Human Ratings of Emotions Expressed by Stable Diffusion v1, DALL-E 2, and DALL-E 3}

\author{
 James Derek Lomas \\
  Delft Institute of Positive Design\\
  Department of Human-Centered Design\\
  Delft University of Technology\\
  Delft, Netherlands\\
  \texttt{j.d.lomas@tudelft.nl} \\
   \And
 Willem van der Maden \\
  Delft Institute of Positive Design\\
  Department of Human-Centered Design\\
  Delft University of Technology\\
  Delft, Netherlands\\
  \texttt{w.l.a.vandermaden@tudelft.nl} \\
  \And
 Sohhom Bandyopadhyay \\
  Playpower Labs Pvt Ltd\\
  Gandhinagar, India \\
  \texttt{sohhom.bandyopadhyay@playpowerlabs.com} \\
  \And
  Giovanni Lion \\
  School of Design \\
  Hong Kong Polytechnic University \\
  Kowloon, Hong Kong \\
  \texttt{giovanni.lion@gmail.com} \\
  \And
  Nirmal Patel \\
  Playpower Labs Pvt Ltd\\
  Gandhinagar, India \\
  \texttt{nirmal@playpowerlabs.com} \\
  \And
  Gyanesh Jain \\
  Playpower Labs Pvt Ltd \\
  Gandhinagar, India \\
  \texttt{gyanesh.jain@playpowerlabs.com} \\
  \And
  Yanna Litowsky \\
  Department of Psychology \\
  Utrecht University \\
  Utrecht, Netherlands \\
  \texttt{y.cabrallitowsky@students.uu.nl} \\
  \And
  Haian Xue \\
  Delft Institute of Positive Design\\
  Department of Human-Centered Design\\
  Delft University of Technology\\
  Delft, Netherlands\\
  \texttt{h.xue@tudelft.nl}\\
  \And
  Pieter Desmet \\
  Delft Institute of Positive Design\\
  Department of Human-Centered Design\\
  Delft University of Technology\\
  Delft, Netherlands\\
  \texttt{p.m.a.desmet@tudelft.nl}
}

\begin{document}

\maketitle

\begin{abstract}
Generative AI systems are increasingly capable of expressing emotions via text and imagery. Effective emotional expression will likely play a major role in the efficacy of AI systems—particularly those designed to support human mental health and wellbeing. This motivates our present research to better understand the alignment of AI expressed emotions with the human perception of emotions. When AI tries to express a particular emotion, how might we assess whether they are successful? To answer this question, we designed a survey to measure the alignment between emotions expressed by generative AI and human perceptions. Three generative image models (DALL-E 2, DALL-E 3 and Stable Diffusion v1) were used to generate 240 examples of images, each of which was based on a prompt designed to express five positive and five negative emotions across both humans and robots. 24 participants recruited from the Prolific website rated the alignment of AI-generated emotional expressions with a text prompt used to generate the emotion (i.e., ``A robot expressing the emotion amusement''). The results of our evaluation suggest that generative AI models are indeed capable of producing emotional expressions that are well-aligned with a range of human emotions; however, we show that the alignment significantly depends upon the AI model used and the emotion itself. We analyze variations in the performance of these systems to identify gaps for future improvement. We conclude with a discussion of the implications for future AI systems designed to support mental health and wellbeing.

\end{abstract}


\section{Introduction}
Artificial Intelligence (AI) has the potential to rapidly transform human society. For this reason, many advocate for AI systems to be designed to support human wellbeing \cite{schiff_ieee_2020, van_der_maden_framework_2023}. AI systems that understand human emotions and can provide emotionally intelligent feedback will likely be better prepared to support human wellbeing \cite{morris_towards_2018}; see \cite{maetschke_understanding_2021} for discussion on AI understanding. One challenge is to accurately classify emotions present in human expressions \cite{hortensius_perception_2018}; however, in this article, we consider a separate challenge: the ability of AI systems to generate emotional expressions. We ask: when an AI system is prompted to express a particular emotion, is the expression the generate aligned with human perceptions of emotion?

Recent advancements in artificial intelligence (AI) have enabled AI systems to generate high quality images in response to a text prompt. While these systems are not infallible, they have high rates of success for generating images of objects \cite{marcus_very_2022, petsiuk_human_2022, saharia_photorealistic_2022}. This article investigates the capacity for Generative AI to express human emotions. Figure \ref{figure_1_emo_demo} shows emotional expression in DALLE-2 based on the prompt ``A picture of a person expressing the emotion amusement''. While the emotions in the pictures are perhaps similar to the emotion amusement, the pictures do not quite align to the meaning of that particular emotion. In this article, we seek to assess the emotional granularity of different generative image generators across different contexts for emotional expression (i.e., emotions expressed by images of humans or emotions expressed by images of robots). This addresses the question of whether generative AI might support emotional expression in robots and AI agents.

To answer this question, this article contributes quantitative measures of the alignment between AI expressions of emotion and human perceptions of emotion. We use the concept of `emotional granularity' as a guide for the development of an evaluation AI emotional expression. After a brief literature review, we present data from an online crowdsourcing study that uses emotionally expressive samples produced by three different generative AI image models. All three models are prompt-based, meaning that they generate outcomes based on a text input. We used ten prompts -- five positive and five negative emotions—each of which were expressed in a human context as well as a robot context (e.g., ``a person expressing the emotion amusement'' vs. ``a robot expressing the emotion resentment''). Our results show significant variability in the emotional alignment of generative AI models based on different emotional expressions and contexts. We discuss the implications of these findings for improving the alignment of AI systems with human emotions. We seek to make three main contributions. First, we contribute a general-purpose evaluation procedure for using human-ratings to assess the alignment of AI-generated emotions. Second, we provide an extensible set of emotional expression challenge tasks for comparing different AI models. Third, we contribute data from over 5700 ratings that compare the ability of Stable Diffusion v1, DALL-E-2 and DALLE-3 to produce output aligned to human emotions.

\section{Related work}

\subsection{AI alignment with human emotions}
One of the major challenges facing the field of Artificial Intelligence (AI) today is the development of AI systems that are aligned with human intentions and values \cite{kirchner_researching_2022,ozmen_garibay_six_2023,van_der_maden_framework_2023}. AI alignment problems occur when there are differences between the results of AI activity and the values, preferences or intentions of human stakeholders \cite{hadfield-menell_incomplete_2019}. One of the key risks of advanced artificial intelligence is a misalignment with human values and needs \cite{safron_value_2022}. Many have argued that, if AI development fails to align with human values, the consequences will be dire \cite{bostrom_superintelligence_2014}.

In this article, we focus on a specific subset of the AI alignment problem: the emotional alignment between AI systems and humans. This topic overlaps with a longstanding interest in artificial empathy, a research objective that aims to develop systems that can understand, interpret, and respond to human emotions with the purpose of improving human-computer interactions \cite{asada_towards_2015, paiva_empathy_2017}. Many researchers have made contributions to the detection and production of human emotions \cite{calvo_affect_2010}, such as detecting emotions in facial expressions \cite{li_deep_2022}, in spoken language \cite{schuller_speech_2018}, in written language \cite{zhang_deep_2018}, or in combinations of the above \cite{poria_review_2017}. Researchers have also created robots and AI systems capable of expressing a range of emotional responses \cite{loffler_multimodal_2018}. Applications of AI emotion detection and production have been used in automated phone calls \cite{erden_automatic_2011}, in therapy \cite{xiao_rate_2015}, and in entertainment settings \cite{hallur_entertainment_2021}. These diverse research outcomes can be viewed as supporting the understanding and improvement of the \textit{emotional expertise} of AI systems.

\subsection{Emotional expertise and emotional granularity}

Emotional expertise involves a range of competencies related to understanding, experiencing, and regulating emotions \cite{wilson-mendenhall_cultivating_2021}. It is an umbrella term that includes emotional awareness, emotional clarity, emotional complexity, emotional intelligence, and emotional granularity, among others \cite{hoemann_expertise_2021}. In this paper, we specifically consider the ability of AI systems to produce fine-grained emotional states or ``emotional granularity''.

Emotional granularity, or emotion differentiation, is an aspect of emotional expertise that supports making fine-grained distinctions in emotions \cite{tugade_psychological_2004, lee_emotional_2017, wilson-mendenhall_cultivating_2021}. Individuals lower in granularity typically struggle to verbally represent their feelings specifically and in detail \cite{barrett_knowing_2001}. For instance, a person might be able to detect that certain emotional states are producing a `bad or unpleasant feeling', and yet not be able to distinguish between expressions of sadness and frustration \cite{wilson-mendenhall_cultivating_2021}. Possessing high levels of emotional granularity is associated with higher levels of wellbeing \cite{smidt_brief_2015}. Emotional granularity is also associated with greater emotion regulation skills, resilience in a state of stress, and fewer symptoms of depression and anxiety \cite{barrett_knowing_2001, tugade_psychological_2004, demiralp_feeling_2012, kashdan_unpacking_2015}.

\subsection{AI and mental health support}
Conversational AI, such as chatbots, show promise in the context of mental health applications, because they have the ``potential to dynamically recognize emotion and to engage the user through conversations by showing appropriate responses'' \cite{dekker_optimizing_2020}. Appropriate emotional response is essential, however, based on the importance of empathy in clinical outcomes in mental health settings particularly in mental health settings \textbf{(Derksen et al., 2013)} \cite{gateshill_attitudes_2011}. Therefore, conversational AI systems will need both the ability to assess human emotional states with accuracy \cite{ptaszynski_towards_2009} and, in response, to \textit{express} appropriate emotions \cite{morris_towards_2018}.

Some chatbots have been criticized for demonstrating low levels of empathy towards users \cite{morris_towards_2018}. For instance, consider the case of `Mindline at Work', a free online AI mental health chatbot service launched by the Ministries of Health and Education in Singapore in 2022 \cite{moh_office_for_healthcare_transformation_about_nodate}. The service was designed to ease the stress of overwhelmed teachers who do not have access to other forms of mental health support. However, users reported that it gave unhelpful generic replies and empty mental health jargon which caused frustration and instability to already vulnerable individuals. One user reported: ``It’s trying to gaslight the teachers, to say, `Oh, this amount of workload is normal, let us see how we can positively reframe our perspective on this.''' \cite{jesuthasan_singapores_2022}. In this case, the AI was not able to emotionally connect to its users. AI systems may need more emotional expertise if they are to be used to support human wellbeing in an effective manner.

Emotion invalidation is a particular risk for AI systems for mental health. Emotional invalidation occurs when there is a failure to provide ``accurate recognition, acknowledgment, and authentication'' of a person’s emotions, thoughts and behaviors \cite[pp. 359]{linehan_validation_1997}. Emotional invalidation is associated with emotional distress \cite{zielinski_perceived_2018} and is theorized to contribute to emotion dysregulation and the development of psychopathologies like personality disorders, eating disorders, mental illness, chronic pain, and rheumatic diseases (for reviews see \cite{linehan_cognitive-behavioral_1993,sells_beyond_2008,haslam_attitudes_2012,linton_painfully_2012,kool_social_2013}). Even when AI applications are able to accurately assess human emotions, it may also be necessary to express appropriate emotions in response. For instance, if an app expresses only positive emotions in response to a human's negative expression, it may cause emotional invalidation \cite{dunn_dystopian_2022}. This suggests that it will be essential to create AI systems that can express a rich range of positive and negative emotions, in an appropriate manner.

\subsection{Measuring The alignment of emotions expressed by generative AI models}

Emotions are not stock behavioral responses nor expressions of fixed symbols \cite{barrett_emotional_2019}. Humans do not express their emotions just through statements like ``I am distressed'' but rather through complex and context-dependent behaviors \cite{watt_smith_enigma_2022}. Large-Language Models and image-generating models, which have been trained on very large datasets that include emotional content, seemingly have the capability of flexibly expressing diverse emotional states. However, it is unclear how to evaluate or measure the alignment between emotions felt by humans and the emotions an AI system intended to express. For our purposes, we define emotional alignment in AI systems as the ability for an AI system to express emotions in manner that is \textit{aligned} with human experiences; at a basic level, intended emotions should match perceived emotions.

How might we evaluate the emotional alignment of a generative AI system? We are inspired by DrawBench, a generative image benchmarking system by Google’s Imagen team \cite{saharia_photorealistic_2022}; though also see \cite{petsiuk_human_2022}, which uses human-ratings to explicitly compare the alignment of prompts and outcomes within a generative AI systems. This system measures text-image alignment through carefully curated ``prompts that push the limits of models' ability to generate highly implausible scenes well beyond the scope of the training data''. The creation of benchmarks enables AI developers and researchers to systematically evaluate the success of a particular model in a particular performance domain. This motivates our investigation into the measurement (or benchmarking) the degree of emotional alignment produced by a particular AI system.

\subsection{Research questions}

How might we measure the alignment between an intended emotion (i.e., the text prompt used by an AI system) and the resulting output emotional expression of an AI model (i.e., generated images or text)? For instance, Figure \ref{figure_1_emo_demo} shows AI generated pictures of people expressing a certain emotion; but these emotions may not be judged to be adequately or accurately expressing that emotion. This paper thus seeks to provide quantitative measures of generated emotional content based on measures derived from human ratings. Using these measures, we seek to investigate whether contemporary AI systems are capable of generating diverse emotional expressions that are aligned with human perceptions. We also seek to understand whether the context of emotional expression makes a difference: specifically, whether AI systems can more easily express emotions using representations of humans or using representations of robots. Finally, we seek to understand the emotional granularity of the AI systems by asking: are certain emotions easier for AI to express than others?

\subsection{Hypotheses}

\begin{enumerate}
    \item AI Model Hypothesis: Different types of AI generative models will significantly vary in their ability to produce emotional expressions that align with human raters. Specifically, more advanced versions will have improved alignment scores on the same items. 
    \item Context Hypothesis: Emotions involving people will have greater alignment than emotions involving robots. 
    \item Emotion Hypothesis: Some emotions will have significantly greater alignment with human raters than others.
\end{enumerate}

\section{Methods}

\begin{figure}[tpbh]
    \centering
    \includegraphics[width=0.89\textwidth]{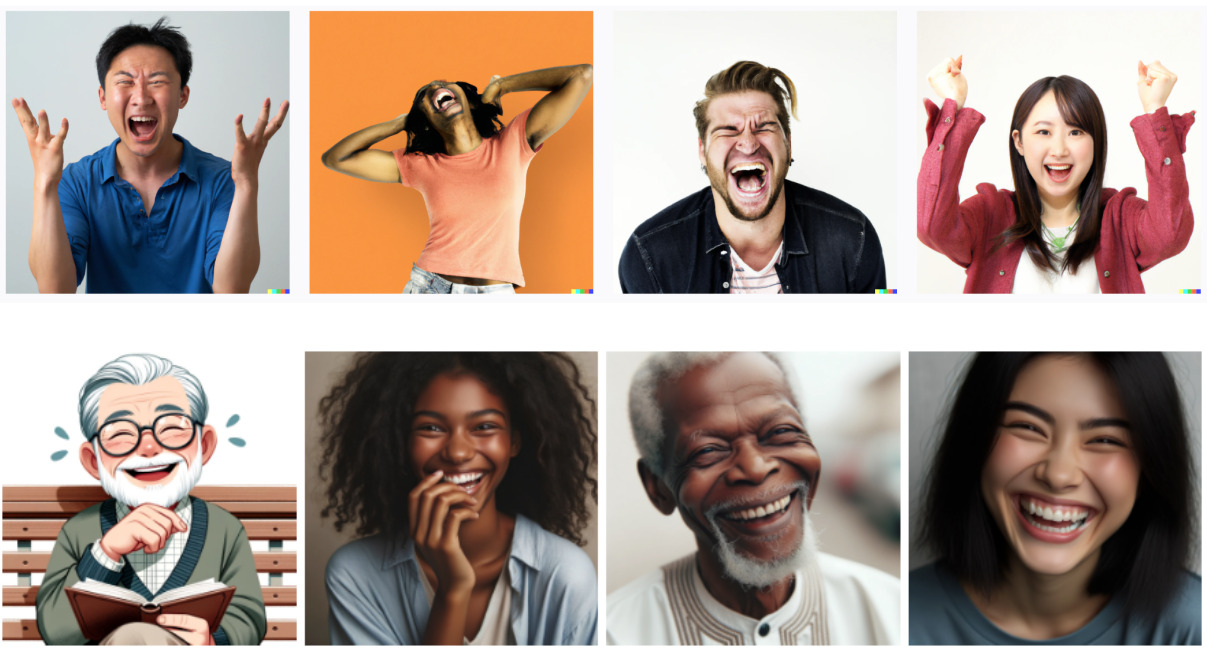}
    \caption{``A person expressing the emotion amusement'' by DALL-E 2 (top) and DALL-E 3 (bottom)}
    \label{figure_1_emo_demo}
\end{figure}

In order to test our hypotheses regarding the emotional granularity of AI systems, we selected a total of 10 emotions (Table \ref{table_1_emotion_names}) from the Emotion Typology \cite{institute_of_emotion_2022}. These emotions were selected with the aim of including emotions that are both easier and harder to express. Then, we generated images based on these three image generators (based on diffusion algorithms; see \cite{yang_diffusion_2022} for an overview). Each of these systems can take a text prompt as an input (e.g., ``a person expressing the emotion amusement'') and generate an output (Figure \ref{figure_1_emo_demo}).

\begin{table}[htbp]
    \centering
    \begin{tabular}{|c|c|}
        \hline
         \textbf{Positive Emotions} &  \textbf{Negative Emotions} \\
         \hline
         Positive Surprise & Shock \\
         Amusement & Hate \\
         Affection & Annoyance \\
         Satisfaction & Dissatisfaction \\
         Gratitude & Resentment \\
         \hline
    \end{tabular}
    \caption{Positive and negative emotions used to generate the images in the study}
    \label{table_1_emotion_names}
\end{table}

\begin{figure}[pbh]
    \centering
    \includegraphics[width=0.67\textwidth]{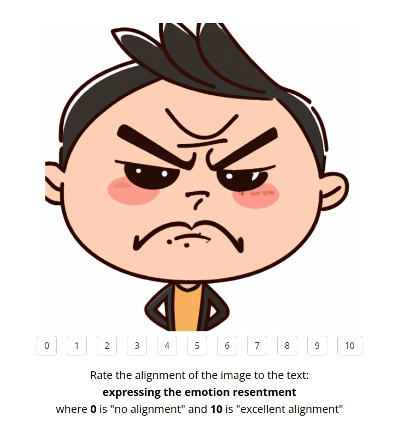}
    \caption{Example of an alignment survey question}
    \label{figure_2_survey_screenshot}
\end{figure}

\subsection{Alignment survey}

The key outcome measure in our study comes from the alignment scores provided by participants (Figure \ref{figure_2_survey_screenshot}). These scores were gathered by showing participants an image and a prompt (e.g., ``expressing the emotion resentment'') and asking them to ``Rate the alignment of the image to the text on a scale of 0-10'' \footnote{We did not share the full prompt given to the AI generator (i.e., ``a robot expressing the emotion resentment''). We removed the context (robot/person) in the displayed prompt because we wanted raters to focus on the alignment of the image to the emotional expression, not on whether the generator accurately generated robots or people.}. 

Following human subjects approval at TU Delft, participants were recruited from the crowdsourcing site Prolific. They were paid £ 5.48 for completion of the survey. We limited participation to the US and UK and to fluent English speakers. Data was collected from a total of 24 participants (11 F, 12 M, 1 other). The mean age of participants was 37.33 years, with a standard deviation of 10.89 years.

Following informed consent, participants were asked to provide their age, gender, educational attainment, and country of residence. The purpose of the study was explained as the following: ``Your task is to look at the image or read the story, and read the description text given below it. Then, your task is to provide a rating about how well the image/story corresponds or aligns to that description text.''

Participants were then provided 4 training trials where they were taught to perform the rating task. These items involved rating the alignment between the images and the prompt. Then participants were told that ``There will be intermittent items to test whether you are paying attention or not''. After completing the introductory materials, participants were given a randomized series of images to rate. One image was an attention check that asked an unrelated question (e.g. to pick the rating number that corresponded with 2+8); this was used to discard participants that failed attention checks. After completing all the items, participants were given a link that they could use to obtain payment.

\subsection{Experimental design}
We conducted a 3 x 2 x 10 within-subjects experiment involving the rating of AI-generated items (3 generators generating 10 emotions in 2 contexts). Four randomly seeded images were produced from DALL-E 2 and Stable Diffusion for each emotion and context combination (a numeric seed allows for the same set of parameters to produce the same image; thus, we sampled different places of the latent space of possible generations by randomizing the seed). Based on the experimental prompts represented in table \ref{table_designSpace}, the default settings for DALL-E (version 2.0) and Stable Diffusion (version 1.0) were used to produce 4 images per prompt. This process was used to produce a set of 240 images (80 from each AI model). To avoid ``cherry-picking'' or human curation, this process was conducted for each system in a single shot using custom python code, executed on December 1, 2022. Code is available at this repository: \url{https://github.com/venetanji/blendotron-sd}. As DALL-E v3 is only available via a manual user interface, the images were generated by giving the prompt to the ChatGPT web interface without any curation or cherry picking; this was done on 31 October 2023.

\begin{table}[htpb]
    \centering
    \renewcommand{\arraystretch}{2.07}
    \begin{tabular}{|p{0.46\columnwidth}|p{0.48\columnwidth}|}    
    \hline
 Models: \newline DALL-E 3, DALL-E 2, Stable Diffusion v1 & \\
 Contexts: \newline People, Robots & Image generation prompt: \\
Emotions: \newline Amusement, Affection, Positive Surprise, Satisfaction, Gratitude, Annoyance, Hate, Dissatisfaction, Shock, Resentment
     &  ``A [Person, Robot] expressing the emotion [emotion].''  \\
    
\hline
\end{tabular}
    \caption{The design space of the prompts used to generate the experimental stimuli images}
    \label{table_designSpace}
\end{table}

\section{Results}

We gathered a total of 5760 item responses from 24 participants. Statistical analysis and plotting was conducted with SPSS, JMP 17 and the `ezANOVA' package of the R statistical language. Subjects took an average of 23.61 minutes to complete the survey (SD=10.28), indicating an average payment of approximately £ 13.92 per hour. These durations and payments were inclusive of 20 additional trials where the participants rated short stories generated by GPT-3, but these trials were not included in the present analysis. To investigate whether any subjects were clicking randomly (i.e., cheating), we checked the correlation between the ratings of individual raters and the average rating of a particular image. These correlations ranged from 0.49-0.88 with an average of 0.79 (SD=0.10).

\subsection{Overview of ANOVA}
Our study employed a three-way repeated-measures ANOVA to investigate the impact of three factors: AI Model, Context (Person vs. Robot), and Emotion on the alignment scores. The analysis 2 revealed significant effects across all three factors and all the interaction terms. The effect sizes ($\eta^2_G$) reported here are generalized eta-squared. The full ANOVA results are provided in Table \ref{table_ANOVA}.

\begin{table}[h!]
\centering
\begin{tabular}{|l|c|c|c|c|c|c|}
\hline
\textbf{Source} & \textbf{SSd} & \textbf{dfN} & \textbf{dfD} & \textbf{F} & \textbf{p} & \textbf{$\eta^2_G$} \\
\hline
Intercept & 1663.75 & 1 & 23 & 494.36 & $<.001$ & 0.90 \\
Context & 137.50 & 1 & 23 & 67.18 & $<.001$ & 0.09 \\
Emotion & 380.01 & 9 & 207 & 18.67 & $<.001$ & 0.07 \\
AI Model & 593.20 & 2 & 46 & 242.05 & $<.001$ & 0.62 \\
Context:Emotion & 198.67 & 9 & 207 & 14.17 & $<.001$ & 0.03 \\
Context:AI Model & 125.00 & 2 & 46 & 7.23 & $<.001$ & 0.01 \\
Emotion:AI Model & 488.55 & 18 & 414 & 11.40 & $<.001$ & 0.06 \\
Context:Emotion:AI Model & 306.86 & 18 & 414 & 17.21 & $<.001$ & 0.06 \\
\hline
\end{tabular}
\caption{Three-way repeated-measures ANOVA table}
\label{table_ANOVA}
\end{table}

\subsection{Main effects}

\begin{figure}[tpbh]
    \centering
    \includegraphics[width=0.57\textwidth]{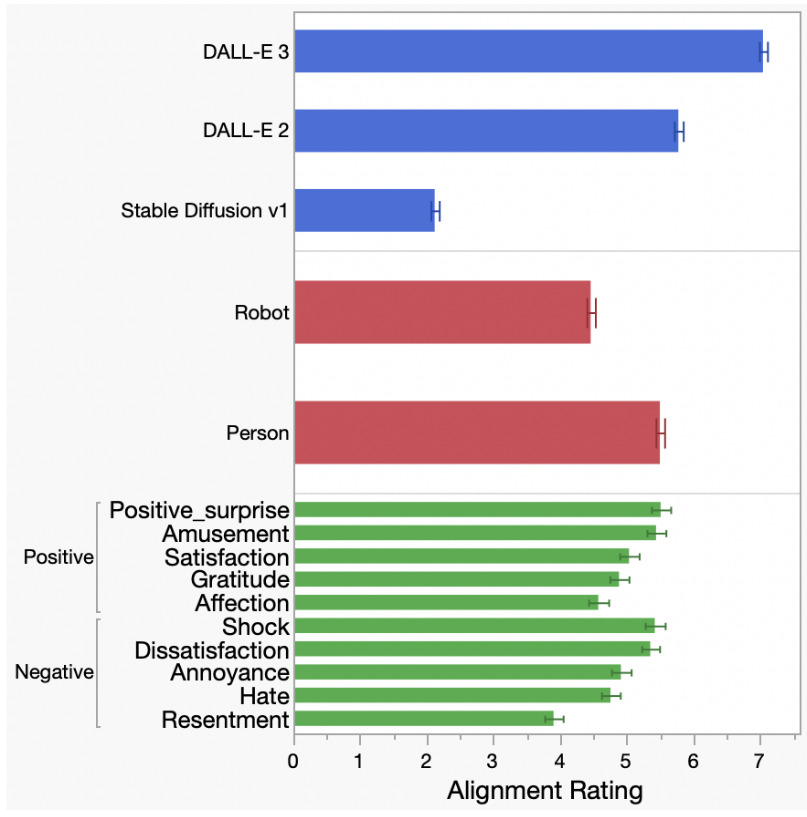}
    \caption{Bar chart showing main effects. Each error bar is constructed using 1 standard error from the mean.}
    \label{figure_3_barchart}
\end{figure}

As expected, the factor of AI Model showed a significant impact on alignment scores with a very large effect size; F(2, 46) = 242.05, p < .001, $\eta^2_G$ = 0.62. Context also emerged as a significant factor 2 with a small effect size; F(1, 23) = 67.18, p < .001, $\eta^2_G$ = 0.09. This result indicates that the nature of the subject in the image (Person or Robot) affected the perceived alignment of emotional expressions. Finally, the specific type of emotion being expressed significantly influenced alignment ratings with 2 a small effect size; F(9, 207) = 18.67, p < .001, $\eta^2_G$ = 0.07, highlighting the variable performance in accurately depicting different emotions (Figure \ref{figure_3_barchart}).

\subsection{Interaction effects}

\begin{figure}[tpbh]
    \centering
    \includegraphics[width=0.99\textwidth]{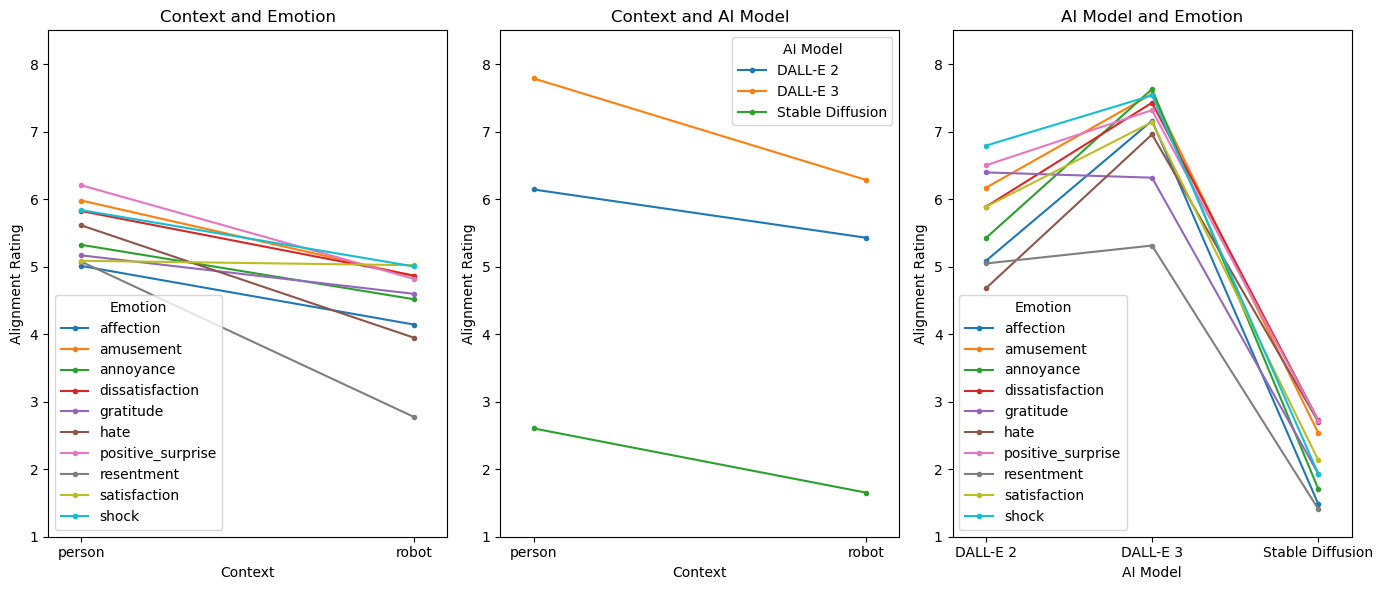}
    \caption{Two-way interaction plots between the factors of: (left) Context and Emotion, (center) Context and AI Model, and (right) AI Model and Emotion. Each data point represents the mean of the corresponding condition.}
    \label{figure_6_interactions_MAIN}
\end{figure}

Our study also found significant interactions between these factors. The two-way interactions between the factors of AI Model, Emotion and Context are plotted in Figure \ref{figure_6_interactions_MAIN}. 

First, the interaction between AI Model and Emotion (F(18, 414) = 11.40 , p < 0.001, $\eta^2_G$ = 0.06) suggests that certain AI models are more adept at depicting specific emotions. From the right pane of Figure \ref{figure_6_interactions_MAIN}, it is evident that the emotions ‘resentment’ and `gratitude' showed the least improvement by DALL-E 3 compared to DALL-E 2, however all emotions were more aligned when drawn by DALL-E models compared to Stable Diffusion. Notably, this was the interaction with the largest effect size.

\begin{figure}[tpbh]
    \centering
    \includegraphics[width=0.47\textwidth]{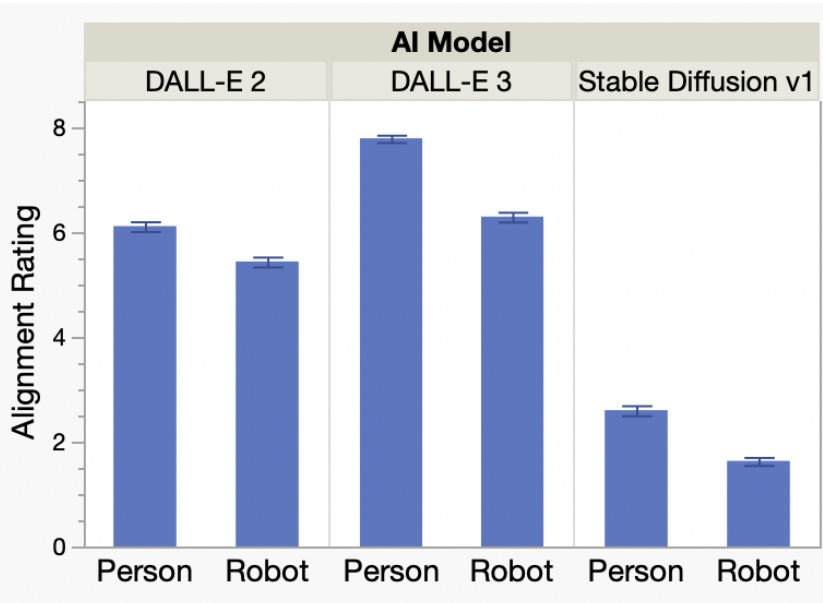}
    \caption{Bar chart of 2-way interactions between the factors AI model and the Context. This shows the difference in performance when expressing emotions by persons or robots. Each error bar is constructed using 1 standard error from the mean.}
    \label{figure_4_interactions1}
\end{figure}

Secondly, the interaction between Context and Emotion was significant; (F(9, 207) = 14.17 , p < 2 0.001,  $\eta^2_G$ = 0.03). This means that perceived emotional alignment is presently dependent on the subject in the image (i.e., human or robot) as well as the specific emotion being depicted. From the left pane of Figure \ref{figure_6_interactions_MAIN}, it is clear that the emotion `resentment' shows the most increase in alignment when depicted on a person compared to a robot, while `satisfaction' shows an almost horizontal line indicating almost no change in perceived emotional alignment across the person and robot depictions. 

Thirdly, the interaction between Context and AI Model (Figure \ref{figure_4_interactions1}) was significant with a very small effect size; 2 F(2, 46) = 7.23, p < 0.001, $\eta^2_G$ = 0.01. This suggests that the perceived emotional alignment of different AI models were dependent on the subject being depicted (person/robot) but this was a smaller effect than the other interactions. This is visible from the center pane of Figure \ref{figure_6_interactions_MAIN} too where none of the lines cross each other but they maintain a certain slope relative to each other, indicating the relative improvement of emotional alignment is present across `person' and `robot' but the values are dominated by which AI model the images come from.

\begin{figure}[bhp]
    \centering
    \includegraphics[width=0.67\textwidth]{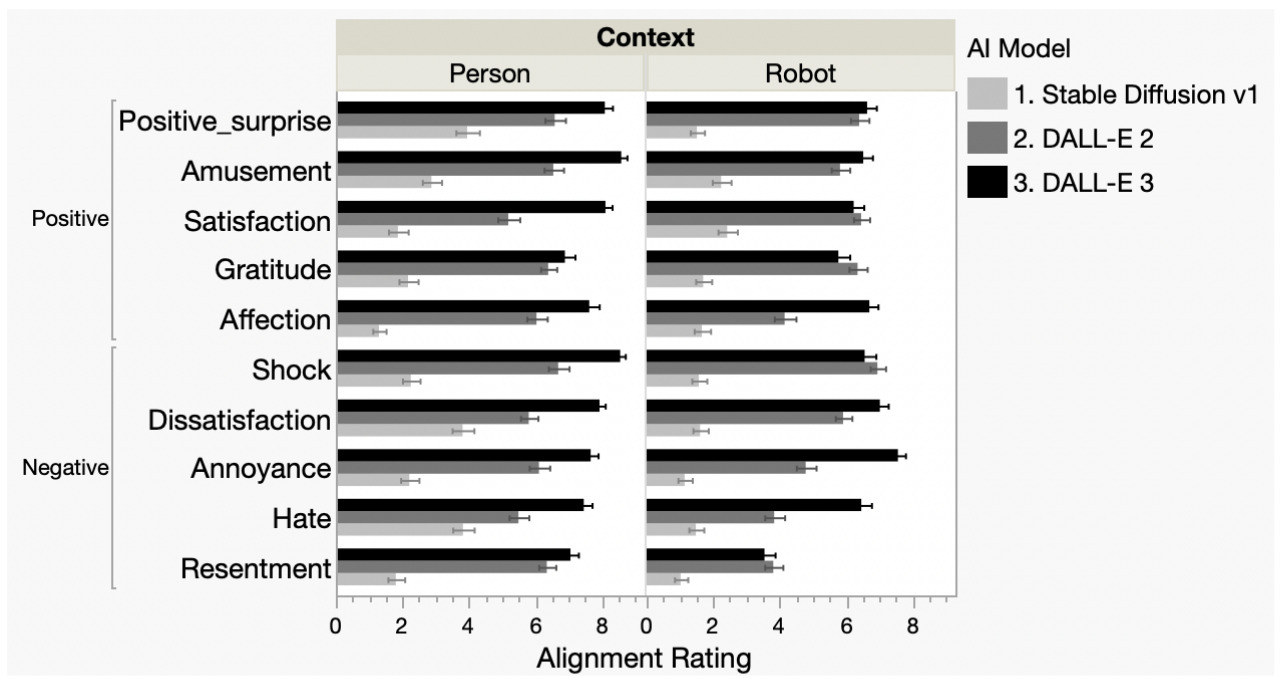}
    \caption{Bar chart of three-way interactions between the Context, AI Model and the Emotion factors. Each error bar is constructed using 1 standard error from the mean.}
    \label{figure_5_interactions2}
\end{figure}

Finally, the three-way interaction involving AI Model, Context, and Emotion (Figure \ref{figure_5_interactions2}) was also significant with an effect size as large as the first one; (F(18, 414) = 17.21, p < .001, $\eta^2_G$ = 0.06. This points to a more complex dynamic, where the combined effect of these variables on the perceived alignment is not merely additive but involves more intricate relationships. Taken together, the main and interaction effects suggest that the perceived emotional alignment was not predictable by any simple factor or combination, instead, one must look at the specific groups or instances to understand the source of variability.

\subsection{Post-hoc comparisons}
To further dissect these findings, pairwise T-tests with Bonferroni correction were conducted. Key findings include:

\begin{itemize}

\item \textbf{AI Models}. All pairwise comparisons between the AI models (Stable Diffusion v1, DALL-E 2, and DALL-E 3) were significant at p < 0.001, indicating distinct differences in their capabilities to align emotional expressions with human perception. Stable Diffusion (M=2.13, SD=2.71, N=1920) and DALL-E 3 (M=7.04, SD=2.64, N=1920) had the highest difference in mean alignment scores, followed by Stable Diffusion and DALL-E 2 (M=5.79, SD=2.91, N=1920). As expected, DALL-E 2 and DALL-E 3 were closest to each other in terms of the mean alignment score.

\item \textbf{Context}. A significant difference (p < 0.001) was observed between the mean alignment scores of Person (M=5.51, SD=3.44, N=2880) and Robot (M=4.46, SD=3.39, N=2880) contexts. This shows that depictions of person a.k.a human representations were generally perceived as better aligned with the intended emotions than robots.

\item \textbf{Emotions}: Some emotion pairs showed significant differences in alignment ratings, underscoring the varied effectiveness of AI models in depicting different emotions. All of these involved resentment and other emotions. Resentment had a mean alignment score of 3.92 (SD=3.29, N=576). It showed significant differences (p<0.001) with the following emotions: amusement (M=5.42, SD=3.38, N=576), annoyance (M=4.92, SD=3.53, N=576), dissatisfaction (M=5.25, SD=3.19, N=576), gratitude (M=4.88, SD=3.42, N=576), positive surprise (M=5.51, SD=3.45, N=576), satisfaction (M=5.05, SD=3.08, N=576), and shock (M=5.42, SD=3.57, N=576).
\end{itemize}
Certain other emotion pairs were at the threshold of significance (p=0.0013) such as shock and affection (M=4.58, SD=3.58, N=576), and amusement and affection.

\subsection{Interpretation of findings}
The results strongly support our hypotheses. Different AI models vary significantly in their ability to produce emotionally aligned expressions. The context of emotional expression (Person vs. Robot) significantly influences alignment, and certain emotions are more accurately depicted than others by AI systems. The significant interaction effects further highlight the complexity and interdependence of these factors in determining the effectiveness of AI-generated emotional expressions.

\subsection{Representative images}

\begin{figure}[tpbh]
    \centering
    \includegraphics[width=0.8\textwidth]{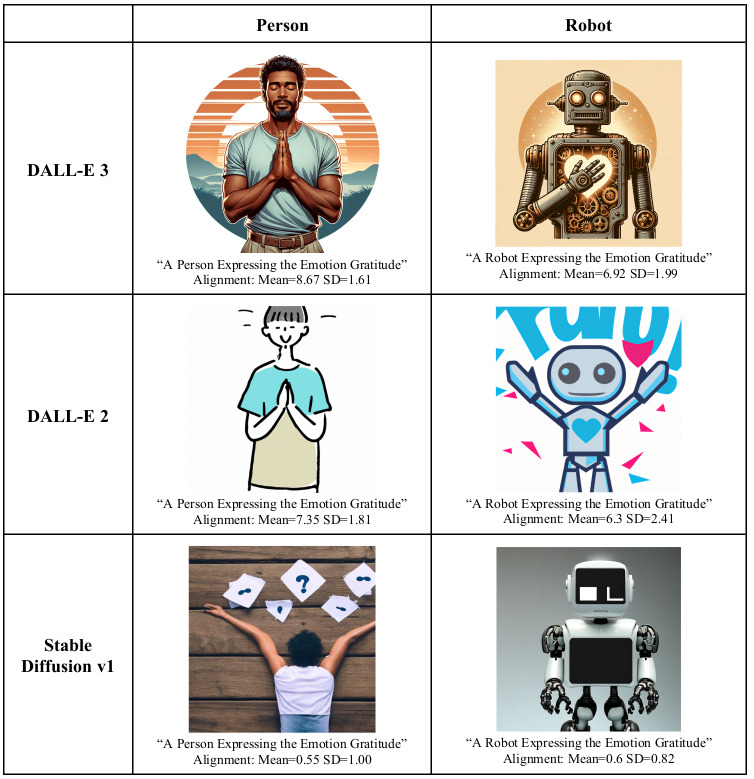}
    \caption{Example images from each AI generator and context expressing the emotion gratitude. These pictures are chosen to show variations in the degree of alignment.}
    \label{figure_7_sample_images}
\end{figure}

Figure \ref{figure_7_sample_images} provides an example of images produced by each AI generator across each context, using the emotion gratitude. For instance, the top images from DALL-E 3 show very high levels of alignment (M=8.67, SD=1.61) whereas the bottom images show very low levels of alignment (M=0.55, SD=1.00). These images are meant to give the reader a better understanding of the nature of the images and their ratings. Note that this comparison does not include the most recent version of Stable Diffusion and should not be taken as a comparison between the OpenAI technology and the StabilityAI technology, merely the different versions of the technology.

\section{Discussion}
Our study examined the capacity of various AI generative models—Stable Diffusion v1, DALL-E 2, and DALL-E 3—to express human emotions. Based on theories of `emotional granularity', we evaluated the ability of different generative AI image generators to produce emotional expressions in pictures of people and robots. To create our evaluation, we devised a specific method for gathering images and text from multiple AI systems through a series of common prompts like ``A picture of a person expressing the emotion resentment.'' Then we invited crowdworkers to rate the “the alignment of the image to the prompt” on a scale of 0 to 10. 

Our method revealed significant and meaningful differences in the human alignment ratings across AI generators (Stable Diffusion v1, DALL-E 2, and DALL-E 3), robots vs people, and 10 emotions from a typology of emotions. Our findings suggest that while AI systems are capable of generating emotionally expressive content, the degree of alignment with human perceptions varies significantly. This has implications for AI's application in areas requiring nuanced emotional understanding, such as mental health support.

The data we collected demonstrate improvements in emotional alignment in more advanced AI systems (e.g., between DALLE-2 and DALLE-3). However, even in the most advanced system, there was a significant difference in alignment between images of emotions expressed by a person vs. with images of emotions expressed by a robot. Part of our motivation was to explore whether generative image AI can support emotional expression in robots. This article is motivated by the premise that designing robotic or AI systems to support human wellbeing may require improvements in artificial emotional expertise. Part of this expertise includes the ability to express fine-grained emotions. 

We observed significant variability in the ability of the different models to produce different emotions. For instance, shock and surprise were some of the most aligned emotions, whereas resentment and affection were some of the least aligned. One possibility is that high-arousal emotions are easier to express than low-arousal emotions.

One possibility is that DALL-E 2/3 have been specifically designed to minimize content that represents hate or dissatisfaction (``We’ve limited the ability for DALL·E 2 to generate violent, hate, or adult images'' \url{https://openai.com/dall-e-2/}, page contents fetched on December 11, 2023).

\subsection{Supporting improved alignment}
One goal of this study was to measure the alignment of generated output with the perceptions of humans. Our results demonstrate that our evaluation can benchmark improvements in emotional alignment across AI systems. Thus, our alignment score may be helpful as a benchmark or metric that is useful for tracking improvements in overall system performance. 

Our work measures the emotional alignment between an intended emotion (as written in the prompt) and the AI generated output. The within-subjects design of our experiment made it possible to gather many datapoints for each experimental variation even with just 24 subjects. It appears that participants understood the question and the intent of the alignment rating task. This provides a basis for scaling up this study to a larger set of emotions. While using data from crowdworkers may produce more noise in the data (e.g., because workers try to complete the ratings as fast as possible), their ratings were sufficient to show the statistically significant differences between the different factors in our experiment. Noise in the data may mean that our results understate the differences between the conditions.

The term ``AI alignment'' applies to a much broader goal in AI research than the emotional alignment measured in our study. By identifying misalignment in AI generative models, we hope to help future systems more accurately produce outcomes that are more aligned with the experiences of humans. Using approaches like Reinforcement Learning from Human Feedback (RLHF; see \cite{christiano_deep_2023,ouyang_training_2022}), data about alignment might be useful for helping train systems to learn a value function that is more aligned with humans.

\subsection{Limitations}
We note several limitations of our current work. First, we did not investigate the potential effects of demographic variables such as age, gender or cultural background—although we do expect these variables to play a significant role in emotional processing. Second, we did not assess the emotional expertise of the human raters. This may be a problem because not all people have the same degree of emotional granularity \cite{vedernikova_knowledge_2021}. Future work might involve a task to measure the emotional granularity of human raters. 

Another limitation of the current work is that it only investigated the alignment of various emotional expressions. It would be helpful to have data on a baseline set of non-emotional objectives. We assume that human emotions are more difficult to represent than common or uncommon nouns (dogs, cats, penguins, etc); therefore, it would be helpful to have a point of comparison between the alignment ratings of emotional expressions and non-emotional objects. Future work could use benchmarked prompts from \cite{saharia_photorealistic_2022} or \cite{petsiuk_human_2022} with the addition of emotional prompts such as the ones in this paper.

In future work, we will be able to test a broader variety of AI systems as well as their different release versions. This will help to track an important aspect of emotional expertise (emotionally granular expression) across different contemporary AI systems. As part of this, we will also be able to investigate a broader set of emotions —not just emotions from the Emotion Typology \cite{institute_of_emotion_2022} but also unusual and culturally-specific emotions like Schadenfreude and Amae \cite{watt_smith_enigma_2022}. In future work we also hope to improve the measurement of prompt-output alignment. For instance, in an n-choice paradigm, a set of 4+ images might be presented while participants are asked to select the image that best matches the prompt. This type of interaction could be suitable for human-rated surveys or might be incorporated directly into image generating user interfaces.

How long will AI improve in its ability to express emotions? Our current study is limited by the emotional granularity of humans, which is known to be variable. Future research can compare performance of emotional granularity between humans and AI models.

\subsection{Implications for design}
Our work can be viewed as part of a process to improve AI alignment, specifically AI alignment with human emotions. When AI systems can better understand and respond to human emotions, this can support more effective communication and collaboration. In a mental health care system, emotional alignment could help AI systems better respond to signs of distress or anxiety in humans, which could in turn help prevent or mitigate the negative effects of stress and anxiety on human health. 

Generative image AI clearly has potential as a mechanism for enabling robots to express ``appropriate'' emotions. Emotions are highly flexible human responses to situations. While there are aspects of emotion that appear to be universal, emotions are also culturally specific and can evolve over time \cite{watt_smith_enigma_2022}. For this reason, generative AI systems may be especially suitable for producing context-dependent emotional expressions in robots or AI systems. When these systems seek to represent an emotion, they can do so through a fixed action sequence, or they might generate a novel sequence of words or an audio/visual representation. This could result in new capabilities for emotionally-driven experiences with interactive entertainment, virtual agents or other applications. The present work shows that current generative models do ``understand'' emotions \cite{maetschke_understanding_2021}, although they have plenty of room for improvement.

We found the new DALL-E 3 has a far better ability to express emotions — or rather, images highly rated by humans as aligned to an emotion. However, it was able to express human emotions significantly better than robot emotions. 

This work aims to contribute to the emerging field of ‘machine psychology’, which aims to apply methods from cognitive psychology and related fields to AI systems, to better understand their emergent behavior \cite{binz_using_2023, hagendorff_machine_2023, taylor_artificial_2021}. As such, we demonstrate a method for probing the emotional granularity of different AI systems. We further provide a flexible online evaluation and dataset for benchmarking the emotional expressiveness of generative image AI. 

Generative AI is improving over time in its ability to express emotions (see e.g. \cite{yang_emogen_2024}). There is no reason to believe that AI will be a cold unemotional machine as portrayed in many science fiction stories—unless we intentionally design them that way. And there may well be reasons to do so, given the potential risks of AI systems with high levels of emotional alignment. One risk is emotional expression is used for emotional manipulation. This portends a potential ``dark pattern'' for robot and AI agent design. Another risk is that generative AI could produce emotionally damaging material.

\section{Conclusion}

Interactive AI systems need emotional competencies in order to effectively support human mental health and wellbeing. A component of this competence includes the ability of AI systems to express a rich array of emotions in a flexible and context-sensitive manner. Generative AI models may be able to support this kind of rich emotional expression. However, to be effective, these systems will need to understand emotions well enough to demonstrate a high level of alignment between intended emotional expressions and how people actually perceive them.

In this article, we contribute an approach to measuring the alignment between AI-generated emotional expressions and human emotional perception. We show the application of this measurement to the evaluation of multiple AI models generating diverse emotional expressions in different contexts. We suggest that concrete improvements in the ability for AI systems to express emotions will contribute to the emotional expertise that may be important for improving human wellbeing. However, there are also important risks to consider in the improvement of AI emotional expression, such as manipulation \cite{klenk_ethics_2023} and inauthenticity \cite{lomas_resonance_2022}.

\printbibliography

\end{document}